\pdfoutput=1
\documentclass[11pt]{article}

\usepackage{acl}

\usepackage{times}
\usepackage{latexsym}

\usepackage[T1]{fontenc}
\usepackage[utf8]{inputenc}

\usepackage{microtype}

\usepackage{inconsolata}

\usepackage{graphicx}
\usepackage{xcolor}
\usepackage{amsmath}
\usepackage{booktabs}
\usepackage{multirow}
\usepackage{cleveref}
\usepackage{enumitem}
\usepackage{amssymb}
\usepackage{stfloats}
\usepackage{tcolorbox}
\usepackage{algorithm}
\usepackage{algorithmic}

\title{Hierarchical Text Classification with LLM-Refined Taxonomies}

\author{
    Jonas Golde\textsuperscript{1,*} \quad Nicolaas Jedema\textsuperscript{2} \quad Ravi Krishnan\textsuperscript{3,*} \quad \textbf{Phong Le\textsuperscript{4,*}} \\
    \textsuperscript{1}Humboldt Universität zu Berlin \quad \textsuperscript{2}Amazon \quad \textsuperscript{3}Meta \\\textsuperscript{4}School of Computer Science, University of St Andrews \\
}

\newcommand{\methodname}[1]{\textsc{TaxMorph}}

\begin{document}
\maketitle
\begin{abstract}
Hierarchical text classification (HTC) depends on taxonomies that organize labels into structured hierarchies. However, many real-world taxonomies introduce ambiguities, such as identical leaf names under similar parent nodes, which prevent language models (LMs) from learning clear decision boundaries. In this paper, we present \methodname{}, a framework that uses large language models (LLMs) to transform entire taxonomies through operations such as renaming, merging, splitting, and reordering. Unlike prior work, our method revises the full hierarchy to better match the semantics encoded by LMs. Experiments across three HTC benchmarks show that LLM-refined taxonomies consistently outperform human-curated ones in various settings up to +2.9pp. in F1.

To better understand these improvements, we compare how well LMs can assign leaf nodes to parent nodes and vice versa across human-curated and LLM-refined taxonomies. We find that human-curated taxonomies lead to more easily separable clusters in embedding space. However, the LLM-refined taxonomies align more closely with the model's actual confusion patterns during classification. In other words, even though they are harder to separate, they better reflect the model’s inductive biases. These findings suggest that LLM-guided refinement creates taxonomies that are more compatible with how models learn, improving HTC performance.
\end{abstract}

\section{Introduction}

\begingroup
\renewcommand\thefootnote{\*}
\footnotetext{*Work done while working at Amazon.}
\endgroup

\begin{figure}[t]
  \centering
  \includegraphics[width=0.95\columnwidth]{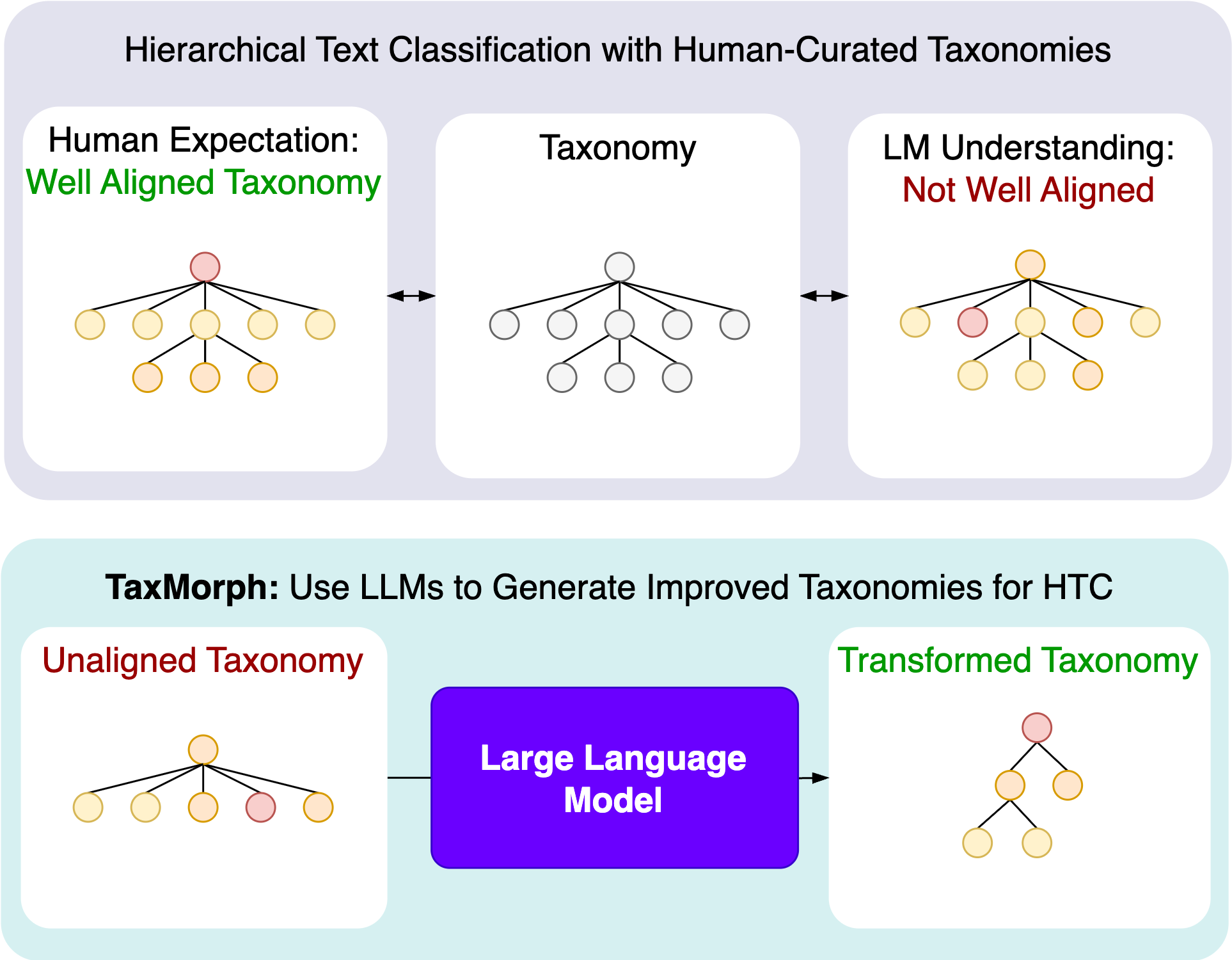}
  \caption{Human-curated taxonomies used in hierarchical text classification (HTC) are often suboptimal due to inconsistencies or ambiguities. \methodname{} transforms these taxonomies into structures that better align with the internal representations of language models, leading to improved performance in HTC tasks.}
  \label{figure:problem_illustration}
\end{figure}

Hierarchies and taxonomies are central to human cognition and language understanding. Organizing information into nested categories enables efficient storage, retrieval, and reasoning about complex concepts \citep{dhillon2002enhanced,sun2001hierarchical}. In natural language processing, hierarchical structures manifest in taxonomies, ontologies, and knowledge bases \citep{vrandecic2014wikidata,lehmann2015dbpedia,heist2021caligraph}.

Hierarchical text classification (HTC) requires models to assign documents to labels organized in a hierarchy. Recent studies have shown that large language models (LLMs) acquire rich hierarchical knowledge during pretraining \citep{lin-ng-2022-bert,wu-etal-2023-plms}. However, their effectiveness in HTC depends on how well their internal representations align with human-curated taxonomies. These taxonomies are manually constructed, often contain redundancies or ambiguities, and may not reflect how LLMs naturally organize or differentiate categories.

For instance, taxonomies may include duplicated or overly generic nodes (e.g., “Design” appearing under both “Web” and “Fashion”) or conflicting class boundaries, which can prevent models from forming clear decision boundaries. Misalignment between the model's representations and the structure of human-defined hierarchies can introduce classification errors and limit performance in HTC.

In this paper, we study whether the structure of a given taxonomy is actually learnable by a language model, and how its structure can be improved to make categories more separable. We introduce \methodname{}, a two-step framework that transforms an existing taxonomy by renaming, merging, splitting, or reordering nodes using an LLM. Unlike prior work, our method operates over the full structure, effectively treating LLMs as taxonomists that revise entire hierarchies in a context-aware manner.

\begin{figure*}[!htb]]
  \centering
  \includegraphics[width=0.95\linewidth]{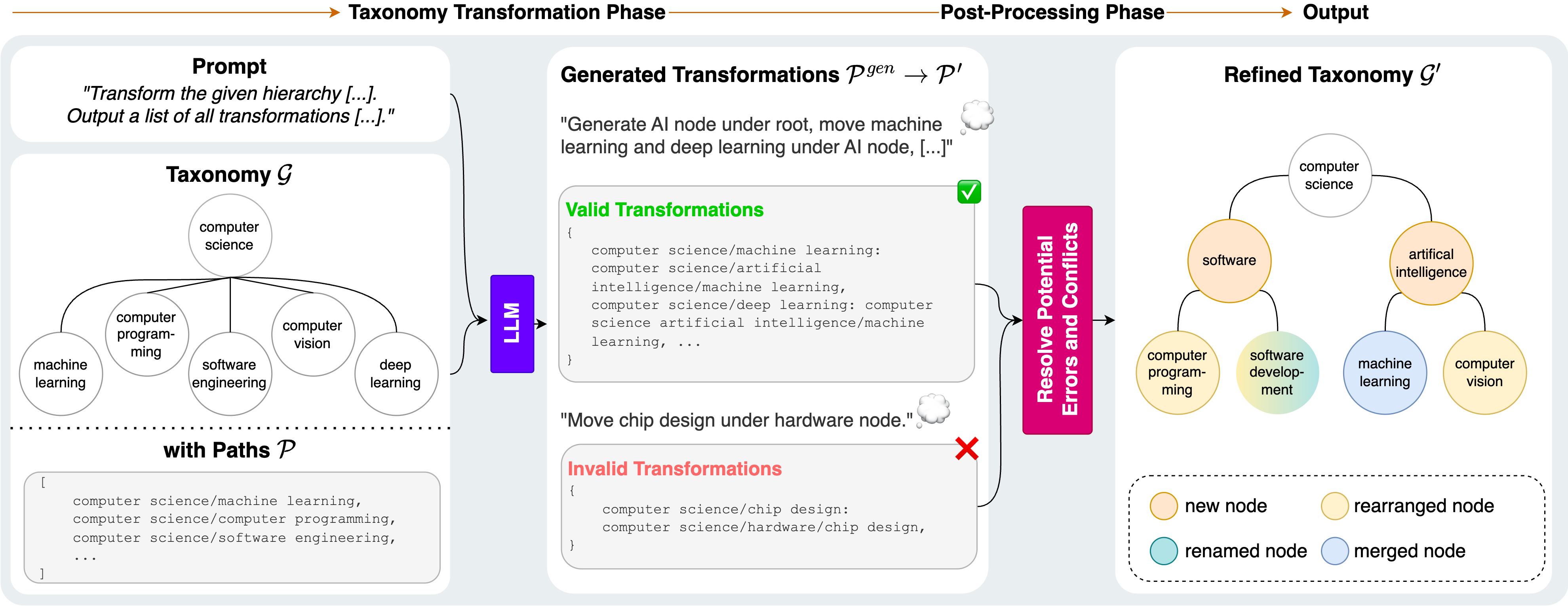}
  \caption {Overview of \methodname{}, a two-step framework for refining taxonomies. In the first phase, an LLM generates transformations by considering the full context of the input taxonomy $\mathcal{G}$. In the second phase, we apply post-processing to correct hallucinations and resolve inconsistencies, resulting in the final refined taxonomy $\mathcal{G}'$.}
  \label{figure:approach}
\end{figure*}

We evaluate our approach across three HTC datasets and show that LLM-refined taxonomies consistently improve classification performance over their human-curated counterparts. To better understand these gains, we analyze the separability of refined hierarchies in embedding space and find that LLM-generated structures better reflect model confusion patterns. These findings highlight the importance of taxonomy structure and suggest that LLMs can play a key role in improving them.

We summarize our contributions as follows:
\begin{itemize}
    \item We propose \methodname{}, a framework for transforming taxonomies using LLMs to improve hierarchical text classification.
    \item We show that modeling with LLM-refined taxonomies consistently outperforms human-curated baselines across multiple datasets.
    \item We conduct detailed ablations to explain why refined hierarchies are more effective, showing they produce harder but more meaningful distinctions in embedding space.
\end{itemize}

\section{\methodname{}} \label{sec:approach}

Creating effective taxonomies involves subtle decisions about granularity, naming, and consistency. These decisions are typically guided by domain knowledge and involve trade-offs similar to those in defining annotation schemes for tasks like named entity recognition (NER) \citep{wang-etal-2019-crossweigh,rucker-akbik-2023-cleanconll}. However, taxonomies introduce additional challenges, as they require structuring an entire concept space rather than classifying isolated entities \citep{kozareva-hovy-2010-semi,bansal-etal-2014-structured}. This often leads to ambiguous or inconsistent label structures, making it difficult for models to learn separable categories \citep{he2024languagemodelshierarchyencoders}.

We propose \methodname{}, a framework that utilizes LLMs as taxonomists by letting the LLM transform and refine a given hierarchy into a more coherent and semantically meaningful structure. \methodname{} consists of two stages: (\textit{1}) a generation phase that produces transformations across the full hierarchy and (\textit{2}) a post-processing phase that resolves inconsistencies in the generated output.

\subsection{Problem Setup}

We focus on hierarchical text classification (HTC), where each input $\mathcal{X}_i \in \mathcal{X}$ is annotated with a subset $\mathcal{Y}_i \subset \mathcal{Y}$ of labels structured in a taxonomy $\mathcal{G} = (V, E)$. Here, $V$ denotes the set of nodes (labels), and $E$ the set of directed edges representing parent-child relationships. We assume $\mathcal{G}$ is a directed acyclic graph (DAG).

Each taxonomy can also be represented as a set of root-to-leaf paths. For each node $v \in V$, we define a unique path $P_v = (v_r, \dots, v)$ from some root node $v_r$ to $v$, where each consecutive pair of nodes is connected by an edge in $E$. The entire taxonomy $\mathcal{G}$ can therefore be expressed as the set of all such paths: $\mathcal{P} = \{ P_v \mid v \in V \}$.

\subsection{Stage 1: LLM-Guided Taxonomy Transformation} \label{sec:generation_phase}

In the first phase (cf.~\Cref{figure:approach}, left), we use an LLM to generate a set of transformations over the original taxonomy $\mathcal{G}$. Concretely, we define a function $f$ that maps each original path $P_v \in \mathcal{P}$ to a JSON object of the form \texttt{\{original\_path: new\_path\}}. To encourage holistic refinements, we prompt the LLM to process the entire taxonomy at once, rather than transforming each node individually. This full-context generation allows the model to propose more coherent restructurings, such as consistent renaming or merging across branches. However, generating the hierarchy in this way also requires keeping track of the original paths (the keys in the JSON object, denoted as $\mathcal{P}^{\text{gen}}$), as they serve as anchors for aligning the transformed paths $\mathcal{P}'$ with the original taxonomy. By comparing each original path with its transformed counterpart, we can identify the changes applied by the LLM. We categorize these transformations into four types:

\begin{itemize}
  \item \textbf{Renamed:} A node's label is semantically adjusted to improve clarity or specificity.
  \item \textbf{Rearranged:} A node is reassigned to a different parent, reflecting a revised understanding of conceptual structure.
  \item \textbf{Generated:} New intermediate nodes are inserted into a path to enrich context or disambiguate meaning. We do not generate new leaf nodes.
  \item \textbf{Merged:} Multiple nodes with overlapping semantics are merged into a single representative node.
\end{itemize}

\subsection{Stage 2: Post-Processing and Error Correction}

In the second stage (cf.~\Cref{figure:approach}, right), we post-process the JSON mapping generated by the LLM, where each key-value pair represents a transformation from an original path $P_v^{\text{gen}}$ to a refined path $P_v'$. Ideally, each key $P_v^{\text{gen}}$ corresponds exactly to an existing path $P_v$ in the original taxonomy $\mathcal{P}$. However, LLM outputs may contain hallucinations or minor deviations such as typos, incorrect formatting, or entirely fabricated entries.

To resolve such cases, we perform a matching step between $P_v^{\text{gen}}$ and the true paths in $\mathcal{P}$ using normalized Levenshtein distance \citep{yujian2007normalized}. If a close match exists, a human reviewer verifies and corrects the key. Otherwise, we discard the transformation. This process allows us to preserve valid transformations while filtering out unreliable ones, all without modifying the original training data or requiring reannotation.

The final refined hierarchy $\mathcal{G}'$ is then constructed from the filtered and corrected set of path mappings $\mathcal{P}'$ which we then subsequently use in our experiments.

\section{Experimental Setup}

\paragraph{Datasets.}~We evaluate our method on four commonly-used hierarchical text classification benchmarks:

\begin{itemize}[leftmargin=1em]
    \item \textbf{Amazon}\footnote{\url{https://www.kaggle.com/datasets/kashnitsky/hierarchical-text-classification}}: Product reviews categorized under a hierarchy of product categories.
    \item \textbf{Books} \citep{aly-etal-2019-hierarchical}: English-language book blurbs categorized into genres using a four-level taxonomy.
    \item \textbf{Web of Science (WOS)} \citep{kowsari2017hdltex}: Scientific paper abstracts organized by academic domain and subdomain.
\end{itemize}

\Cref{table:dataset_statistics} provides summary statistics, and \Cref{table:label_per_level} shows the number of labels at each depth level.

\begin{table}[!ht]
\centering
\small
\begin{tabular}{lccccc}
\toprule
Dataset & Classes & Depth & Train & Val & Test \\
\midrule
Amazon & 584 & 3 & 37.7k & 4.7k & 4.7k \\
Books & 152 & 4 & 58.7k & 14.8k & 18.4k \\
WOS & 152 & 2 & 32.9k & 7.0k & 7.0k \\
\bottomrule
\end{tabular}
\caption{Dataset statistics, including total number of labels, maximum taxonomy depth, and split sizes.}
\label{table:dataset_statistics}
\end{table}

\begin{table}[!ht]
\centering
\small
\begin{tabular}{c|cccc}
\toprule
Level & Amazon & Books & WOS \\
\midrule
1 & 6 & 7  & 7 \\
2 & 64 & 52 & 145 \\
3 & 514 & 77 & -- \\
4 & -- & 16 & -- \\
\bottomrule
\end{tabular}
\caption{Number of labels per level in each dataset’s taxonomy.}
\label{table:label_per_level}
\end{table}

\subsection{Experimental Setup} \label{sec:experiments_model}

\paragraph{Taxonomy Refinement with LLMs.}~To refine the taxonomies, we consider three LLMs: \texttt{Haiku}, \texttt{Sonnet-3}, and \texttt{Sonnet-3.5}. All models used have long-context capabilities and are sufficiently large to enable strong zero-shot performance, which is essential given the lack of specialized models for taxonomy refinement. For each model, we generate three independent refinements, following the procedure described in~\Cref{sec:approach}. This allows us to assess both model-dependent and stochastic variability in the refined hierarchies.

\paragraph{Hierarchical Classification Model.}~For downstream evaluation, we use a bi-encoder model that encodes both input texts $\mathcal{X}$ and taxonomy labels $\mathcal{Y}$ using shared parameters $\theta$. Given a text input $\mathcal{X}_i$ and a candidate label $y \in \mathcal{Y}$, we compute their similarity using a dot product:
\[
s = \theta(\mathcal{X}_i)^\top \cdot \theta(y),
\]
where $\theta(\cdot) \in \mathbb{R}^d$. The resulting scores are passed through a sigmoid function to obtain probabilities. We train the model using binary cross-entropy loss, and during inference, we predict a label if its probability exceeds a threshold of 0.5. 

We use DistilBERT~\citep{sanh2020distilbertdistilledversionbert} as the encoder across all experiments. We employ AdamW optimizer~\citep{loshchilov2018decoupled}, with a learning rate of $2 \times 10^{-5}$, a batch size 32, and a maximum of 10{,}000 training steps. All experiments are implemented using Hugging Face Transformers \citep{wolf-etal-2020-transformers} and PyTorch \citep{pytorch2024}.

\paragraph{Label Representations.}
We explore four label representations to assess the role of semantic and structural information in hierarchical classification.

\begin{itemize}[leftmargin=1em]
    \item \textbf{Single Node:} Each label is represented using only its leaf node name (e.g., \texttt{machine learning}), without parent or ancestor context. This setting uses only local semantics.
    \item \textbf{Full Path:} Each label is represented by its full hierarchical path from the root (e.g., \texttt{computer science $\rightarrow$ artificial intelligence $\rightarrow$ machine learning}), providing additional contextual information that can help disambiguate similar leaf nodes.
\end{itemize}

\noindent To better isolate the contribution of label semantics, we also include two control baselines that remove natural language descriptions from the labels.

\begin{itemize}[leftmargin=1em]
    \item \textbf{Linear Layer:} A standard classifier that treats labels as output indices with no associated text or structure. This setup tests whether the model can learn class boundaries without any semantic guidance.
    \item \textbf{Structured Identifier:} Each label is replaced by a code (e.g., \texttt{1.10.20}) that reflects its location in the hierarchy but omits all semantic content. This setting retains hierarchical structure while discarding label meaning.
\end{itemize}

\begin{table*}
\centering
\begin{tabular}{ll|l|llll|l}
\toprule
Model & Dataset & Invalid & Renamed & Rearranged & Generated & Merged & Unchanged \\
\midrule
\multirow[c]{3}{*}{Haiku} & WOS & $20.7_{ 18.1}$ & - & $118.7_{ 14.6}$ & $26.3_{ 8.7}$ & - & $17.0_{ 0.0}$ \\
 & Books & $10.0_{ 9.0}$ & $78.0_{ 22.6}$ & $6.5_{ 1.5}$ & $2.5_{ 0.5}$ & - & $31.0_{ 14.9}$ \\
 & Amazon & $1.5_{ 0.5}$ & $257.0_{ 0.0}$ & $38.0_{ 20.0}$ & $12.5_{ 8.5}$ & $14.0_{ 4.0}$ & $52.0_{ 0.0}$ \\
\midrule
\multirow[c]{3}{*}{Sonnet (3)} & WOS & - & $90.0_{ 4.0}$ & $102.5_{ 2.5}$ & $55.5_{ 2.5}$ & - & $18.5_{ 1.5}$ \\
 & Books & - & $108.0_{ 7.0}$ & - & - & - & $13.5_{ 6.5}$ \\
 & Amazon & $1.0_{ 0.0}$ & $397.0_{ 3.0}$ & - & - & $7.0_{ 0.0}$ & $4.0_{ 0.0}$ \\
\midrule
\multirow[c]{3}{*}{Sonnet (3.5)} & WOS  & - & $126.0_{ 1.0}$ & $138.0_{ 1.0}$ & $82.0_{ 0.0}$ & - & $1.0_{ 0.0}$ \\
 & Books & - & $108.0_{ 0.0}$ & $2.5_{ 0.5}$ & $2.0_{ 0.0}$ & - & $8.5_{ 0.5}$ \\
 & Amazon & - & $36.2_{ 19.6}$ & $1.0_{ 0.0}$ & $1.0_{ 0.0}$ & - & $2.2_{ 1.6}$ \\
\bottomrule
\end{tabular}
\caption{Transformations applied through the LLM to the human-curated taxonomy, categorized by type.}
\label{table:transformation_stats}
\end{table*}

\section{Experiments}

\subsection{Experiment 1: Transformations Applied through the LLM}

We analyze the transformations applied by \methodname{} across datasets and LLMs (cf.~\Cref{table:transformation_stats}). Renaming is the most common change, particularly in deeper or more complex taxonomies. For example, Sonnet-3 renames 397 nodes in the Amazon taxonomy covering over two-thirds of its 584 labels. In WOS, Sonnet-3.5 reassigns 138 nodes to different parents, nearly matching the 145 total labels at level 2, indicating that the LLM reorganizes the majority of the hierarchy. We also observe a substantial number of generated nodes; for instance, Sonnet-3.5 inserts 82 new intermediate nodes in WOS to improve semantic separation between paths.

We further assess the reliability of these refinements by tracking invalid transformations, e.g. cases where the LLM generates paths not present in the original taxonomy. Such errors are rare and mostly confined to Haiku, which produces an average of 21 invalid paths in WOS. In contrast, Sonnet-3.5 yields zero invalid paths across all datasets, highlighting that more advanced LLMs can restructure large, real-world hierarchies with high precision. Overall, the scale and type of transformations correlate with the depth and complexity of the original taxonomy, suggesting that \methodname{} effectively adapts to the structure of its input.

Finally, a human reviewer was consulted in only 0.6\% of all cases, all of which involved minor typos in the LLM-generated paths. In each instance, our string matching procedure correctly identified the intended existing path, demonstrating its effectiveness in resolving such errors automatically.

\subsection{Experiment 2: Performance on HTC}

\Cref{table:main_results} presents the results of fine-tuning on the taxonomies generated by \methodname{}. Across all evaluated benchmarks, we observe consistent improvements in classification performance when using LLM-refined taxonomies. The highest overall performance is achieved by \methodname{} using Sonnet-3.5 in the Single Node setting, attaining an average F1 score of 0.625, compared to 0.608 for the best-performing human-curated baseline. The largest improvement occurs on the Books dataset, where the F1 score increases from 0.583 to 0.612, corresponding to a gain of +2.9 percentage points. On the Amazon dataset, the Haiku-based taxonomy achieves the best performance with an F1 score of 0.475, outperforming the baseline score of 0.457. In the WOS dataset, the gains are smaller but still notable, with Sonnet-3.5 improving the F1 score from 0.785 to 0.802.

In addition to the prompt-based approaches that leverage semantic label descriptions, we also evaluate two control conditions that exclude semantic information. The Linear Layer baseline performs worst overall, with an average F1 score of 0.403, highlighting the difficulty of learning the label space without access to textual descriptions. Using structured identifiers as label representations improves performance to 0.542, but remains substantially below any semantic variant. This result underscores the importance of label semantics for effective learning. Between the two prompt-based settings, the Single Node representation consistently outperforms the Full Path representation (e.g., 0.612 vs. 0.607 for Sonnet-3.5), suggesting that precise, well-aligned leaf-level labels are more informative than full hierarchical paths in the context of \methodname{}. This observation indicates that deeper hierarchical context may introduce redundancy or noise, which can interfere with the model’s ability to learn fine-grained distinctions.

\begin{table*}
\centering
\begin{tabular}{llcccr}
\toprule
Taxonomy & Label Representation & Amazon & Books & WOS & Avg. \\
\midrule
\multirow[c]{4}{*}{Human-Curated} 
  & Linear Layer & 0.142 & 0.348 & 0.720 & 0.403 \\
  & Structured Identifier & 0.351 & 0.508 & 0.767 & 0.542 \\\cline{2-6}
  & Single Node & 0.457 & 0.583 & 0.785 & 0.608 \\
  & Full Path & 0.432 & 0.582 & 0.783 & 0.599 \\
\midrule
\multirow[c]{2}{*}{\methodname{} (Haiku)} 
  & Single Node & \textbf{0.475} & 0.610 & 0.786 & 0.624 \\
  & Full Path & 0.457 & 0.603 & 0.790 & 0.617 \\
\midrule
\multirow[c]{2}{*}{\methodname{} (Sonnet-3)} 
  & Single Node & 0.462 & 0.609 & 0.791 & 0.621 \\
  & Full Path & 0.439 & 0.602 & 0.789 & 0.610 \\
\midrule
\multirow[c]{2}{*}{\methodname{} (Sonnet-3.5)} 
  & Single Node & 0.460 & \textbf{0.612} & \textbf{0.802} & \textbf{0.625} \\
  & Full Path & 0.447 & 0.607 & 0.796 & 0.617 \\
\bottomrule
\end{tabular}
\caption{Results when using \methodname{} with different LLMs. We report macro-averaged F1 scores. The final column reports the average across the three datasets.}
\label{table:main_results}
\end{table*}

\subsection{Experiment 3: Few-Shot Experiments}

\begin{figure*}[t]
   \includegraphics[width=\linewidth]{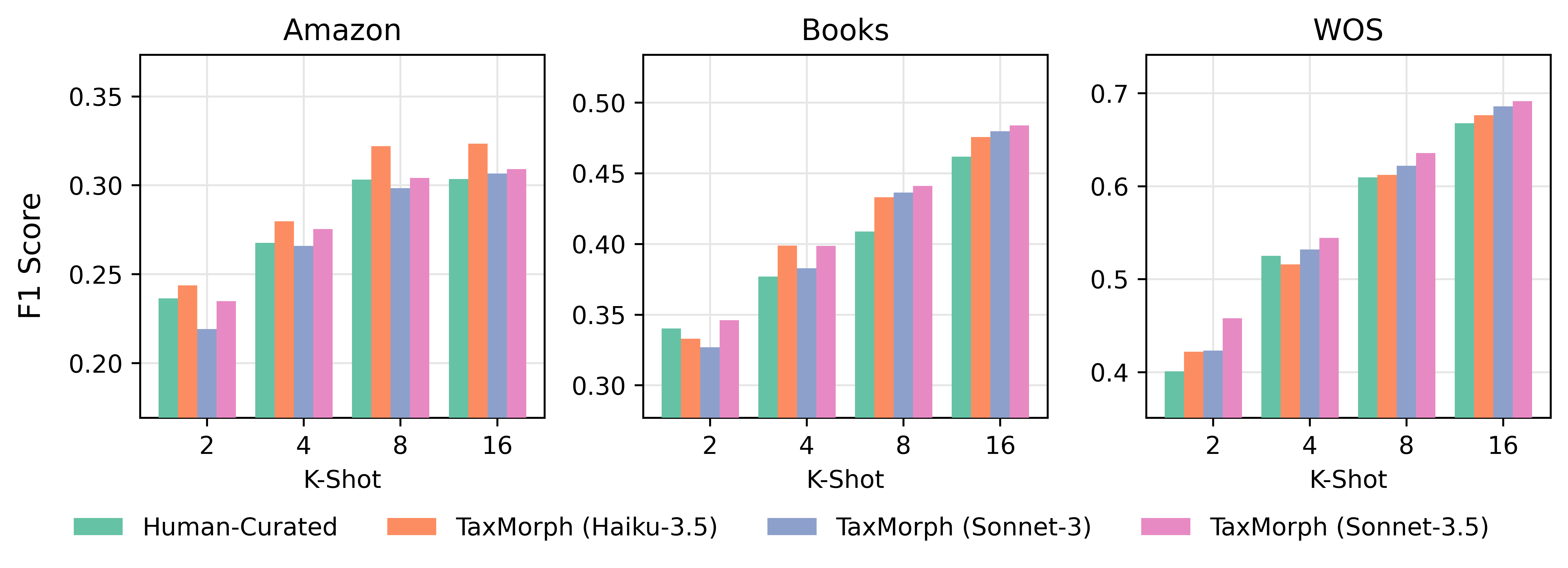}
\caption{Few-shot performance on hierarchical text classification with human-curated versus LLM-refined taxonomies using \methodname{}. We report macro-averaged F1 scores across four datasets using DistilBERT and the Single Node setting, varying the number of training examples per leaf node ($k \in \{2, 4, 8, 16\}$). LLM-refined taxonomies consistently outperform human-curated ones, especially as more training data becomes available.}
   \label{fig:fewshot_results}
\end{figure*}

We evaluate the impact of \methodname{} in few-shot settings by training models with only $k \in \{2, 4, 8, 16\}$ examples per leaf node. As previous experiment shows improvements in full fine-tuning settings, we now investigate if improved taxonomies can also help in data-scarce scenarios. All other hyperparameters and training procedures follow the setup in~\Cref{sec:experiments_model}, with the exception that training is limited to 2{,}000 steps to reduce overfitting under low-resource conditions.

The results, shown in~\Cref{fig:fewshot_results}, demonstrate that taxonomies refined with \methodname{} consistently outperform human-curated taxonomies across all datasets as the number of examples increases. On the Books dataset, the F1 score improves from 0.462 (human-curated) to 0.484 with \methodname{} (Sonnet-3.5) at $k = 16$, representing the largest observed gain in this setting. In the WOS dataset, the best-performing configuration at $k = 16$ is again \methodname{} (Sonnet-3.5), achieving 0.691 compared to 0.667 for the human-curated hierarchy. The Amazon dataset shows smaller improvements overall, though \methodname{} (Haiku-3.5) outperforms all other configurations at $k = 8$ and $k = 16$, reaching 0.322 and 0.323, respectively.

At $k = 2$, differences between methods are less pronounced. In Books and Amazon, the human-curated taxonomy performs comparably or slightly better than some LLM-refined versions. However, in WOS, even under this low-resource condition, \methodname{} (Sonnet-3.5) achieves a substantial improvement, reaching 0.458 versus 0.401 for the human-curated baseline. These results indicate that while the benefit of refinement increases with data availability, LLM-refined taxonomies can already provide gains in the low-shot regime, particularly in settings with well-structured hierarchies.

\subsection{Experiment 4: Embedding Space Analysis}

One central assumption is that a well-structured hierarchy should support the model in assigning input texts to their correct positions within the taxonomy. When this assignment fails, the cause may be either in the intrinsic difficulty of the input or in the taxonomy itself due to vague labels, overlapping categories, or inconsistent structure. In this ablation experiment, we investigate to what extent a language model learns a true hierarchical layout in the embedding space and whether LLM-generated taxonomies are easier learn to learn.

\citet{paletto-etal-2024-label} propose local embedding-based metrics to assess taxonomy coherence, such as measuring the similarity between sibling nodes (to which we refer as child-to-child similarity (CS) subsequently) or averaging the similarity between parents and their children (to which we refer as parent-to-child similarity (PS) subsequently). While these metrics capture surface-level structure, they do not support a deeper understanding whether a language model truely learns a hierarchical structure. To address this, we introduce the Taxonomy Probing Metric (TPM), a probing-based evaluation that directly tests whether the structure of a taxonomy is unambiguously differentiable in embedding space. To do so, TPM evaluates (1) whether each child node is most similar to its correct parent among all candidate parents, and (2) whether each parent is most similar to its correct set of children among all possible child groupings. This formulation mirrors the inference process in classification: the model must choose the correct node among many. Since our setting uses taxonomy-aware prompting, TPM is designed to test whether the hierarchical structure induces meaningful and recoverable clusters in embedding space. Crucially, it provides a way to connect structural alignment to predictive behavior by asking whether the taxonomy actively supports the model in making correct decisions. We provide details on the the metric in~\Cref{sec:metric_definition,sec:tpm}.

\begin{figure*}[ht]
  \centering
  \includegraphics[width=0.95\linewidth]{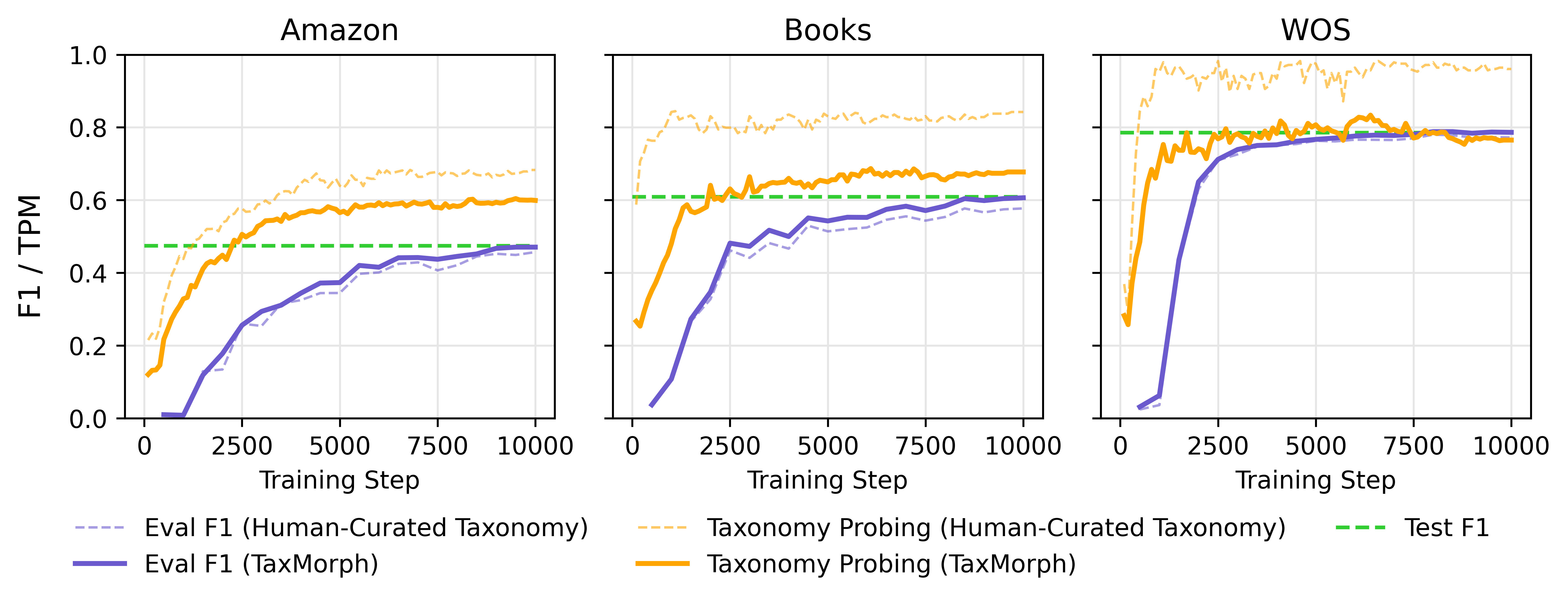}
  \caption{Comparison of F1 and our Taxonomy Probing Metric (TPM) between human-curated taxonomy and LLM-refined ones. We report the metrics for setting Prompt (Single Node) and using Haiku for refining the taxonomy.}
  \label{fig:train_dynamics_main}
\end{figure*}

\begin{figure*}[ht]
  \centering
  \includegraphics[width=0.95\linewidth]{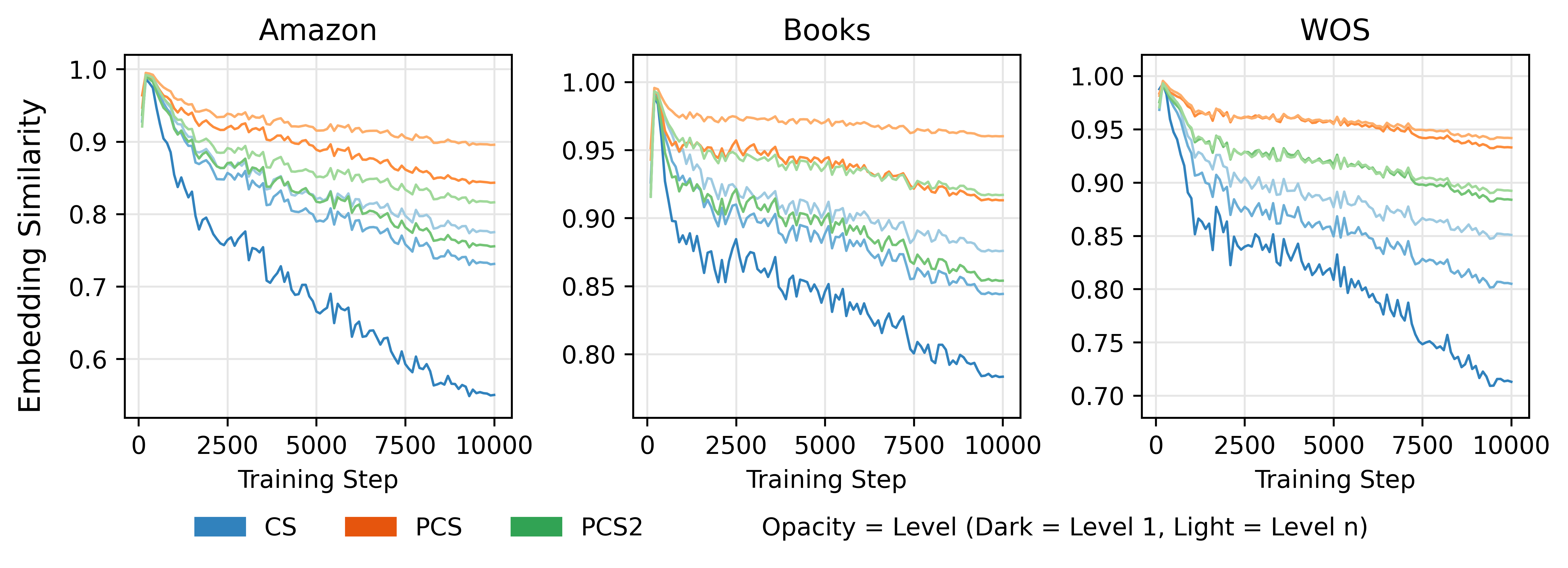}
\caption{Semantic similarity trends of embedding-based taxonomy alignment metrics during fine-tuning. We plot Children Similarity (CS), Parent-Child Similarity (PCS), and Parent-Child Centrality (PCS2) across training steps using \methodname{} with Haiku-3.5 and the Prompt (Single Node) setting. The metrics quantify local coherence in embedding space but do not correlate with downstream classification performance.}
  \label{fig:related_work_metrics}
\end{figure*}

In~\Cref{fig:train_dynamics_main}, we plot (1) the F1 score on the evaluation set during fine-tuning, (2) our proposed TPM metric, both for (3) models fine-tuned on LLM-refined taxonomies generated by TaxMorph (using Haiku-3.5), and (4) models fine-tuned on the original human-curated hierarchy. In~\Cref{fig:related_work_metrics}, we show the alignment metrics introduced by \citet{paletto-etal-2024-label} (Children Similarity (CS), Parent-Child Similarity (PCS), and Parent-Child Centrality (PCS2)) along with our additional metric PS2, which captures the cosine similarity between a parent and the average embedding of its children.

When comparing TPM to CS, PCS, and PCS2, we observe that TPM correlates much more closely with downstream F1 performance. While TPM generally increases as the model learns, the embedding-based metrics from prior work show a consistent decline during training. This trend suggests that the model pushes node representations apart in embedding space, making absolute similarity values less meaningful for evaluating structural alignment. In contrast, TPM directly captures whether the model can differentiate parents and children among competing alternatives, aligning more with the model’s decision-making objective during inference.

We further observe that TPM scores are consistently lower for LLM-refined taxonomies than for their human-curated counterparts. This suggests that while LLM-generated hierarchies lead to improved classification performance, they introduce structural complexity that makes them harder to interpret in embedding space. Qualitative analysis reveals that LLMs tend to enrich taxonomies by inserting intermediate nodes and refining label semantics, resulting in more expressive structures. These refinements improve task performance but do not necessarily yield tighter or more regular geometric organization in the embedding space. This highlights an important distinction: taxonomies that support better predictions may not be those that are easiest to embed or align spatially.

\section{Related Work}
\textbf{Hierarchical Text Classification.}~Existing research leverages the general language understanding capabilities of large LMs (LLMs) for hierarchical text classification (HTC) \citep{zhou-etal-2020-hierarchy,chen-etal-2021-hierarchy,bhambhoria-etal-2023-simple,wang-etal-2023-towards-better}. More recent works have advanced these approaches by incorporating contrastive learning techniques \citep{wang-etal-2022-incorporating,zhu-etal-2024-hill} to better differentiate between similar classes, or by integrating external knowledge from knowledge graphs \citep{liu-etal-2023-enhancing,jain-etal-2024-higen} to enrich the classification process. These methods typically assume that the underlying taxonomy is fixed and well-formed.

Our approach diverges from this assumption by shifting focus from classification architectures to the structure of the taxonomy itself. We treat the taxonomy not as a static input but as an object that can be refined and optimized using LLMs. This allows us to improve compatibility between the label space and the model’s inductive biases, yielding gains in downstream performance even without changing the classifier. Moreover, our evaluation goes beyond task accuracy to assess structural alignment, providing insights into how changes in the hierarchy shape the learning dynamics of HTC.

\textbf{Taxonomy Understanding.}~Beyond HTC, taxonomies play a central role in many areas of natural language processing, including taxonomy induction from unstructured text \citep{bosselut-etal-2019-comet,chen-etal-2021-constructing}, reasoning over transitive and hierarchical relations \citep{aly-etal-2019-every,lin-ng-2022-bert,tran2024training,he-etal-2023-language}, and understanding structured knowledge representations \citep{wu-etal-2023-plms,zhou2024doesmapotofucontain,moskvoretskii-etal-2024-large}. Several recent works investigate whether LMs encode relational or hierarchical structure internally, often using probing tasks or synthetic benchmarks \citep{richard-etal-2024-fracas}. Specialized models have also been proposed that explicitly encode hierarchies, including those trained with structural inductive biases \citep{moskvoretskii-etal-2024-taxollama} or using hyperbolic embeddings to capture the exponential growth of tree-like structures \citep{he2024languagemodelshierarchyencoders}.

While these works aim to better model taxonomic structure, their evaluations often rely on local assessments, such as pairwise relation prediction \citep{petroni-etal-2019-language}, masked token likelihoods, or similarity-based matching \citep{eval-harness,wiland-etal-2024-bear}. Our work complements and extends this literature by proposing a probing-based metric designed to reflect global alignment between model representations and hierarchical structure. Rather than evaluating alignment indirectly through static embeddings or isolated edges, we directly measure whether a given hierarchy helps the model make correct predictions.

\section{Conclusion}
Our study focuses on improving hierarchical text classification by refining human-curated taxonomies using large language models. We introduce \methodname{}, a framework that leverages LLMs to generate more expressive hierarchical structures, leading to consistent performance gains across multiple HTC benchmarks. Beyond performance, we investigate how these LLM-refined taxonomies behave in embedding space. To this end, we propose the Taxonomy Probing Metric (TPM), which directly tests whether parent-child relationships remain distinguishable after fine-tuning. Our findings show that while LLM-generated hierarchies boost classification accuracy, they often reduce structural clarity in embedding space, suggesting a trade-off between expressiveness and geometric alignment. These results underscore the need to evaluate taxonomies not only by task performance but also by how well their structure aligns with model representations.

\clearpage

\section*{Limitations}
Our study investigates the use of LLMs to refine taxonomies for hierarchical text classification. While our \methodname{} approach yields improved downstream performance, it has several limitations:

\begin{enumerate}
\item The structure and content of the refined taxonomies are not yet fully explored. While the evaluation shows improved classification performance, it does not address questions related to semantic validity, potential biases, or alignment with domain-specific expectations. Additional analyses are needed to better understand the nature of the refinements produced by LLMs.

\item Although the experiments include multiple classifiers such as DistilBERT and BERT, the results are restricted to encoder-based transformer models. Other classes of models, including generative architectures, graph-based architectures or those incorporating external knowledge, are not considered. The analysis may therefore not generalize to alternative modeling choices.

\item Additional limitations include the exclusive use of LLMs from a single family (Anthropic Claude), the reliance on fixed prompting templates, and the absence of human evaluation to assess the quality of taxonomy edits. These constraints may limit the interpretability and scope of the findings.

\end{enumerate}

Future work should investigate a broader range of model architectures, refinement strategies, and evaluation protocols to assess the structural properties and practical implications of LLM-generated taxonomies.

\bibliography{custom}

\clearpage
\appendix

\section{Measuring Taxonomy Alignment} \label{sec:metric_definition}

To assess the quality of refined taxonomies beyond downstream task performance, we evaluate how well language models capture the underlying hierarchical structure. We adopt embedding-based metrics proposed by \citet{paletto-etal-2024-label} that quantify local taxonomic coherence in the model's representation space. Let $\theta$ denote a sentence encoder (e.g., a sentence transformer~\citep{reimers-gurevych-2019-sentence}) and let $\varphi(\cdot, \cdot)$ denote cosine similarity.

Given a parent node $v_i$ and its set of direct children, we define $V_i = \{v_i\} \cup \text{children}(v_i)$. The \textbf{Children Similarity (CS)} metric computes the average pairwise cosine similarity among all children of a given parent:

\[
CS(V_i) = \frac{\sum_{v_j, v_k \in \text{children}(v_i)} \varphi (\theta (v_j), \theta (v_k))}{\binom{|\text{children}(v_i)|}{2}}.
\]

The \textbf{Parent-Child Similarity (PCS)} metric measures how close a parent is to each of its children in the embedding space:

\[
PCS(V_i) = \frac{\sum_{v_j \in \text{children}(v_i)} \varphi (\theta (v_i), \theta (v_j))}{|\text{children}(v_i)|}
\]

We further consider the alternative \textbf{Parent-Child Centrality (PCS2)} metric as the similarity between a parent node and the average embedding of its children:

\[
PCS2(V_i) = \varphi\left( \theta(v_i), \overline{\theta}(\text{children}(v_i)) \right)
\]
where
\[
\overline{\theta}(\text{children}(v_i))=\frac{\sum_{v_j \in \text{children}(v_i)} \theta (v_j)}{|\text{children}(v_i)|}.
\]

These metrics quantify how locally consistent a taxonomy is in the embedding space. However, they are limited to pairwise or cluster-level relationships and may not fully capture structural confusion in the global hierarchy. In particular, dense clusters or weakly separated branches can result in misleadingly high scores despite poor alignment with the intended taxonomy structure.

\section{Taxonomy Alignment Metric} \label{sec:tpm} To evaluate a model’s understanding of the full hierarchical structure, we introduce a probing-based approach that measures alignment between nodes in the taxonomy and their embeddings. This method captures global structural coherence rather than local similarity alone. We use the terminology
\begin{itemize}
    \item Let \( V^l \) be the set of nodes at level \( l \) (the children).
    \item Let \( V^{l-1}\) be the set of nodes at level \( l-1 \) (the parents).
    \item Let \( \text{children}(v) \) be the set of children of node $v$.
    \item Let \(\theta(v)\) be the embedding of a node \( v \).
    \item Let \(\varphi(\cdot, \cdot)\) be the cosine similarity function.
\end{itemize}

\noindent\textbf{Top-down matching.}~This probing method evaluates whether the model can correctly associate each parent node with its corresponding set of children. The core assumption is that the average embedding of a set of child nodes should reflect the embedding of their common parent.

Let \( V^l \) denote the set of nodes at level \( l \), and let \( \text{children}(v^l_k) \) denote the set of children for a given parent node \( v^l_k \) at level \( l \). For each parent node \( v^l_i \in V^l \), we compute cosine similarity between its embedding and the average embedding of all child sets at level \( l + 1 \). The model is then tasked with matching each parent to the most similar child set:
\begin{gather*}
M = \{(v^l_i, \text{children}(v_{k^*}^l)) \mid \\ 
k^* = \arg\max_{k} \varphi(\theta(v^l_i), \overline{\theta}(\text{children}(v_k^l) )) \}
\end{gather*}
where \( M \) is the resulting set of predicted parent–child set associations. A prediction is considered correct if the assigned child set exactly matches the true children of the parent. We report top-down accuracy ($TD$) as the proportion of correctly matched pairs in \( M \) across all levels of the hierarchy.

\noindent\textbf{Bottom-up matching.} This probing method complements the previous one by verifying whether each child is assigned to its correct parent. Let $V^{l-1}$ denote the set of parent nodes at level $l-1$, and $V^l$ the children at level $l$. For each parent $v_i^{l-1}$, we compute a decision boundary $D(v_i^{l-1})$, defined as the minimum similarity between the parent and any of its children:
\[
D(v_i^{l-1}) = \min_{v_j \in \text{children}(v_i^{l-1})} \varphi(\theta(v_i^{l-1}), \theta(v_j)).
\]
Given a child $v^l_j \in V^l$, we predict its parent as the node $v_k^{l-1}$ with highest cosine similarity:
\[
k = \arg\max_{v_i^{l-1} \in V^{l-1}} \varphi(\theta(v^l_j), \theta(v_i^{l-1})).
\]
The prediction is counted as correct only if: (1) $v_k^{l-1}$ is the true parent, (2) the similarity exceeds the decision boundary $D(v_k^{l-1})$, and (3) the similarity does not exceed the boundary of any other parent. We denote the resulting accuracy as $BU$.

Finally, we define the Taxonomy Probing Metric as the harmonic mean of top-down and bottom-up matching scores:
\[
\text{TPM}(\mathcal{G}) = \frac{2 \cdot TD \cdot BU}{TD + BU}.
\]
This metric captures both local consistency and global structural understanding of the taxonomy. It rewards models that both preserve the intended hierarchy and avoid ambiguous assignments in the embedding space.

\section{Results Using Different Transformers}

In this section, we present the main results obtained using BERT \citep{devlin-etal-2019-bert}. The experimental setup is identical to the one described in the main section. 

The results are consistent with those in the main section, suggesting that taxonomies refined with \methodname{} are also effective for other student models, including those that are more powerful. However, we encountered some training instabilities with our hyperparameter configuration in the Linear Layers and Simple Enumerations settings, as the results for certain datasets were significantly lower compared to those obtained with DistilBERT.

\begin{table*}
\centering
\begin{tabular}{llcccr}
\toprule
Taxonomy & Label Representation & Amazon & Books & WOS & Avg. \\
\midrule
\multirow[c]{4}{*}{Human-Curated} 
  & Linear Layer & 0.009 & 0.247 & 0.093 & 0.116 \\
  & Structured Identifier & 0.322 & 0.518 & 0.762 & 0.534 \\\cline{2-6}
  & Single Node & 0.455 & 0.582 & 0.783 & 0.607 \\
  & Full Path & 0.457 & 0.594 & 0.778 & 0.610 \\
\midrule
\multirow[c]{2}{*}{\methodname{} (Haiku)} 
  & Single Node & \textbf{0.477} & 0.591 & 0.793 & 0.620 \\
  & Full Path & 0.457 & 0.610 & 0.789 & 0.619 \\
\midrule
\multirow[c]{2}{*}{\methodname{} (Sonnet-3)} 
  & Single Node & 0.466 & 0.611 & \textbf{0.794} & 0.624 \\
  & Full Path & 0.464 & 0.618 & 0.786 & 0.623 \\
\midrule
\multirow[c]{2}{*}{\methodname{} (Sonnet-3.5)} 
  & Single Node & 0.471 & \textbf{0.616} & 0.798 & \textbf{0.628} \\
  & Full Path & 0.467 & 0.618 & 0.793 & 0.626 \\
\bottomrule
\end{tabular}
\caption{Main results when using BERT trained on LLM-refined hierarchies with \methodname{}. We report macro-averaged F1 scores. The final column shows the average across Amazon, Books, and WOS.}
\label{table:main_results_bert}
\end{table*}

\section{Pseudocode for Counting Transformations}

In~\Cref{alg:short_transformation}, we present the pseudocode used to count and classify the transformations applied by \methodname{}. Each original path is compared to its refined counterpart to determine whether it was renamed, rearranged, merged, or extended with new nodes. This classification allows us to analyze how \methodname{} modifies the taxonomy.

\begin{algorithm}[!ht]
\caption{Taxonomy Transformation Classification}
\label{alg:short_transformation}
\begin{algorithmic}[1]
\REQUIRE Mapping $f: \mathcal{P} \rightarrow \mathcal{P'}$ from original to generated paths

\FORALL{$p \in \mathcal{P}$}
    \IF{$p \notin \text{dom}(f)$}
        \STATE \texttt{invalid}
    \ELSE
        \STATE Let $p = (c_1, \ldots, c_n)$, $f(p) = (c'_1, \ldots, c'_m)$

        \IF{$(c_1, \ldots, c_n) = (c'_1, \ldots, c'_m)$}
            \STATE \texttt{unchanged}
        \ELSIF{$n = m$}
            \STATE \texttt{renamed (count each $c_i \neq c'_i$)}
        \ELSIF{$n < m$}
            \STATE \texttt{generated} (+ \texttt{renamed/rearranged} if endpoints differ)
        \ELSIF{$n > m$ or $f$ not injective}
            \STATE \texttt{merged}
        \ENDIF
    \ENDIF
\ENDFOR
\end{algorithmic}
\end{algorithm}

\section{Full Prompt for \methodname{}}

\begin{tcolorbox}[colback=gray!5, colframe=gray!80!black, fonttitle=\bfseries]
You are an expert taxonomist. Your task is to refine a given taxonomy represented as a list of slash-separated hierarchical paths. Each path is a string encoding a sequence of categories from root to leaf, such as \texttt{"electronics/computers/laptops"}.

Your goal is to produce a refined taxonomy by transforming these paths. Valid transformation types include:

\begin{itemize}
  \item \textbf{Renaming} categories (e.g., \texttt{"pet supplies"} $\rightarrow$ \texttt{"pets"})
  \item \textbf{Rearranging} levels (e.g., moving a category under a more appropriate parent)
  \item \textbf{Generating} intermediate categories to increase granularity
  \item \textbf{Merging} semantically redundant categories
\end{itemize}

You are not allowed to generate new, additional leaf nodes. Only modify a path if there is a clear reason. Output must be in \textbf{strict JSON} format: a dictionary mapping each original path to its refined version. Both keys and values must be strings using the same slash-separated format.

\textbf{Input taxonomy (example):}
\begin{verbatim}
[
  "electronics/computers/laptops",
  "grocery/fruit/citrus/orange",
  "grocery/beverages/soft drinks",
  ...
]
\end{verbatim}

\textbf{Expected output format:}
\begin{verbatim}
{
  "electronics/computers/laptops": 
  "electronics/computing/laptops",
  ...
}
\end{verbatim}
\end{tcolorbox}

\section{Expressiveness vs. Alignment}

\begin{figure*}[t]
  \centering
  \includegraphics[width=0.95\linewidth]{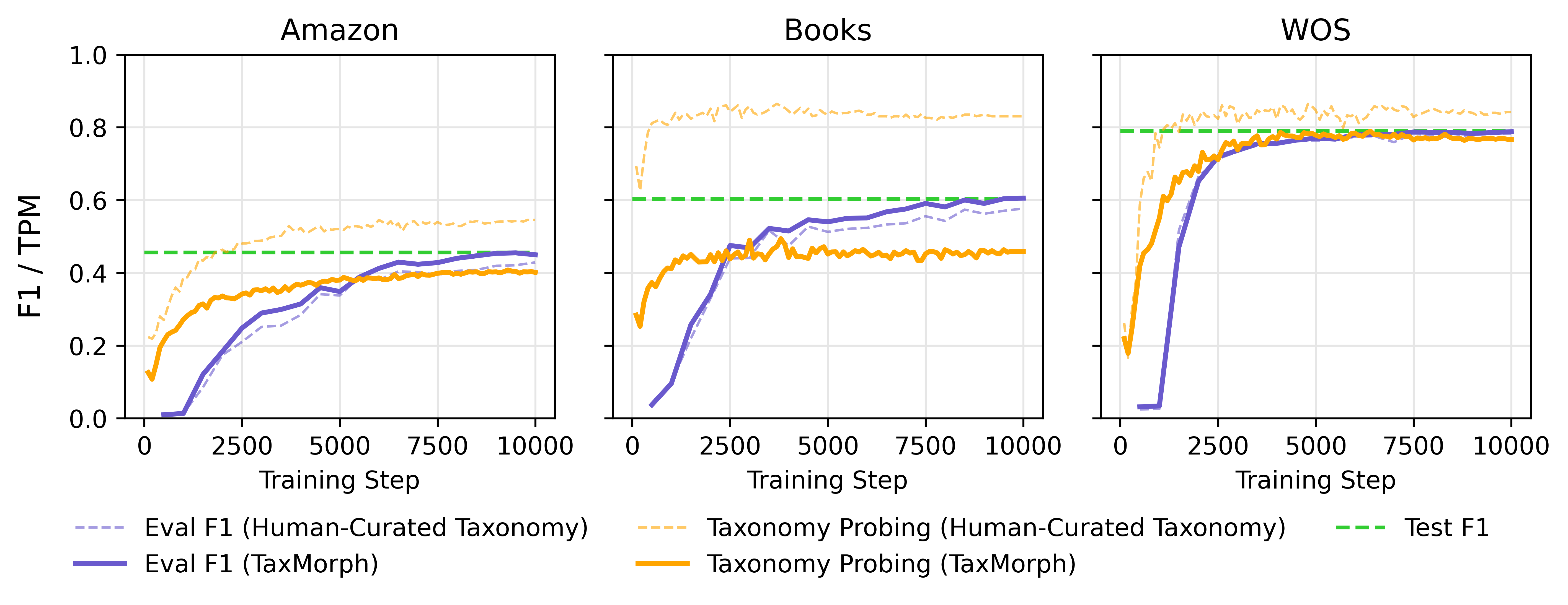}
  \caption{Comparison of F1 and our Taxonomy Probing Metric (TPM) between human-curated taxonomy and LLM-refined ones. We report the metrics using the entire hierarchy path as label semantics. We use Haiku-3.5 for refining the taxonomy.}
  \label{fig:train_dynamic_ablation}
\end{figure*}

In~\Cref{fig:train_dynamic_ablation}, we present the same training dynamics as in~\Cref{fig:train_dynamics_main}, but using more expressive label semantics by providing the full path during fine-tuning. Surprisingly, performance decreases despite the increased expressiveness. Although the model can, in principle, learn parent-child relationships in this setting, it becomes more easily confused. This suggests that taxonomies are best understood by language models when labels at each level are clear and discriminative.

\end{document}